\documentclass{article}
\ifdefined\pdfsuppresswarningpagegroup\pdfsuppresswarningpagegroup=1\fi
\usepackage{iclr2027_conference,times}
\iclrfinalcopy
\usepackage[T1]{fontenc}
\usepackage[utf8]{inputenc}
\usepackage{amsmath,amssymb,booktabs,array,tabularx,graphicx,microtype,xcolor,colortbl}
\usepackage{url,float}
\usepackage{algorithm,algpseudocode}
\usepackage[most]{tcolorbox}
\usepackage{hyperref}

\hypersetup{hypertexnames=false,
  pdftitle={EvoMCP: Co-Evolving MCP Agents and Training Data},
  pdfauthor={Xuanang Gao; Kepeng Lin; Xinyan Shi; Yong Deng; Shuheng Shen; Changhua Meng; Xiaofeng Wu; Jie Yang; Wei Liu}}
\definecolor{oursrow}{HTML}{EAF3FB}
\definecolor{promptblue}{HTML}{245B8A}
\definecolor{promptback}{HTML}{F3F7FB}
\newtcolorbox{promptcard}[1]{
  enhanced,
  nobeforeafter,
  width=0.98\linewidth,
  colback=promptback,
  colframe=promptblue,
  boxrule=0.8pt,
  arc=2mm,
  left=3mm,
  right=3mm,
  top=5mm,
  bottom=2mm,
  title={#1},
  fonttitle=\bfseries,
  coltitle=white,
  colbacktitle=promptblue,
  attach boxed title to top left={xshift=4mm,yshift=-2.5mm},
  boxed title style={arc=1.5mm,boxrule=0pt,left=2mm,right=2mm,top=0.8mm,bottom=0.8mm},
  fontupper=\small,
  before upper={\raggedright\setlength{\parindent}{0pt}\setlength{\parskip}{3pt}}
}
\newcommand{\method}{EvoMCP}
\newcommand{\D}{\mathcal{D}}
\newcolumntype{L}[1]{>{\raggedright\arraybackslash}p{#1}}
\newcolumntype{Y}{>{\raggedright\arraybackslash}X}
\title{\raggedright EvoMCP: Co-Evolving MCP Agents\\and Training Data}
\author{%
  \bfseries Xuanang Gao$^{1,2}$\thanks{Equal contribution. Contact: \href{mailto:fangkuar@sjtu.edu.cn}{\texttt{fangkuar@sjtu.edu.cn}}.},
  Kepeng Lin$^{2,3}$\footnotemark[1],
  Xinyan Shi$^{2,4}$\footnotemark[1],
  Yong Deng$^{2}$,\\
  \bfseries Shuheng Shen$^{2}$,
  Changhua Meng$^{2}$,
  Xiaofeng Wu$^{2}$,
  Jie Yang$^{1}$ \&
  Wei Liu$^{1}$\thanks{Corresponding author.}\\
  \normalfont $^{1}$Shanghai Jiao Tong University\\
  \normalfont $^{2}$Ant Group\\
  \normalfont $^{3}$Huazhong University of Science and Technology\\
  \normalfont $^{4}$Nanjing University of Aeronautics and Astronautics
}

\begin{document}
\maketitle
\fancyhead{}
\renewcommand{\headrulewidth}{0pt}
\addtocontents{toc}{\protect\setcounter{tocdepth}{-1}}

\begin{abstract}
Supervised fine-tuning of Model Context Protocol (MCP) agents is typically conducted on a static data distribution, which fails to adapt to the student's evolving capabilities and leaves newly exposed weaknesses insufficiently addressed.
To address these challenges, we propose \method, an agent--data mutual evolution framework that enables MCP agents to improve through a closed loop of task evolution, trajectory evolution, and supervised fine-tuning.
Specifically, \method\ assesses the student's established capabilities and capability gaps from rollout feedback to identify learning targets.
Task evolution then translates these targets into executable tasks by adapting the instructions and observable environment states of compatible verified seed tasks while keeping their verifiers aligned.
To provide reliable and efficient demonstrations for these tasks, trajectory evolution uses reusable skills to recover failed teacher rollouts and refine successful ones.
The resulting task--trajectory pairs are used to fine-tune the student, while accepted tasks seed subsequent evolution.
To close the loop, we re-evaluate the updated student and use its feedback to guide the next round of data evolution.
Experiments on Toolathlon, MCPMark, and MCP-Atlas demonstrate consistent improvements over the initial student, yielding absolute gains of 16.5\%, 10.2\%, and 12.6\%, respectively.
Project page: \url{https://github.com/xagao/EvoMCP}.
\end{abstract}

\section{Introduction}
\label{sec:introduction}

Large language model (LLM) agents handle multi-step workflows that involve finding suitable tools, invoking them with appropriate parameters, responding to tool failures, and combining their outputs into accurate, context-aware responses \citep{xu2025toucansynthesizing15mtoolagentic}. Recent advances, including the Model Context Protocol (MCP), simplify tool integration through standardized interfaces, connecting LLMs to real-world environments and making it easier for agents to discover, invoke, and execute external tools \citep{xu2025toucansynthesizing15mtoolagentic,wang-etal-2026-mcp}. To develop these capabilities, models are trained on pre-collected trajectories using supervised fine-tuning (SFT) \citep{xu2025toucansynthesizing15mtoolagentic,he-etal-2026-webstar}. An instance of this training data comprises a task--trajectory pair, where trajectories capture sequences of planning, tool calls, tool responses, and the final model response \citep{xu2025toucansynthesizing15mtoolagentic}.

Existing MCP data pipelines synthesize tasks and tool-use trajectories for agent training \citep{xu2025toucansynthesizing15mtoolagentic,wang-etal-2026-mcp,lu2026fireflyilluminatinglargescaleverified,xu2026envfactoryscalingtooluseagents}. While synthetic data reduces reliance on human annotation, a fixed training corpus cannot adapt to the student's evolving capabilities, making it difficult to target specific weaknesses or support continual improvement \citep{yang-etal-2026-coevolve}. For MCP agents trained with SFT, adapting the training data requires executable tasks paired with reliable teacher demonstrations.

Constructing these task--trajectory pairs requires keeping the task instruction, environment, and verifier mutually consistent \citep{li2026recursivesynthesislonghorizonterminal,song2026envscalerscalingtoolinteractiveenvironments}. For example, an MCP agent cannot detect declining quiz scores without access to current and historical records. The task instruction and verifier must also agree on which declines should be flagged. Once these tasks are validated, teacher rollouts provide demonstrations for student training. Training on these demonstrations resembles hard distillation in language model training, where a teacher model generates outputs and a student model learns to imitate them \citep{he-etal-2026-webstar}. A teacher may fail to complete an executable task. However, even successful teacher rollouts can contain incorrect or suboptimal actions \citep{he-etal-2026-webstar}. This underscores the need for more principled strategies to transform raw rollout trajectories into high-quality training data.

To address these challenges, we propose \method, an agent--data mutual evolution framework that enables MCP agents to improve through a closed loop of task evolution, trajectory evolution, and supervised fine-tuning. Our core idea is to use student rollout feedback to assess established capabilities and capability gaps, identify learning targets, and guide task evolution accordingly. We therefore evolve complete task--trajectory pairs, adapting compatible verified seed tasks to these targets and improving the teacher demonstrations used to train the student.

\begin{figure}[t]
\centering
\includegraphics[width=\textwidth]{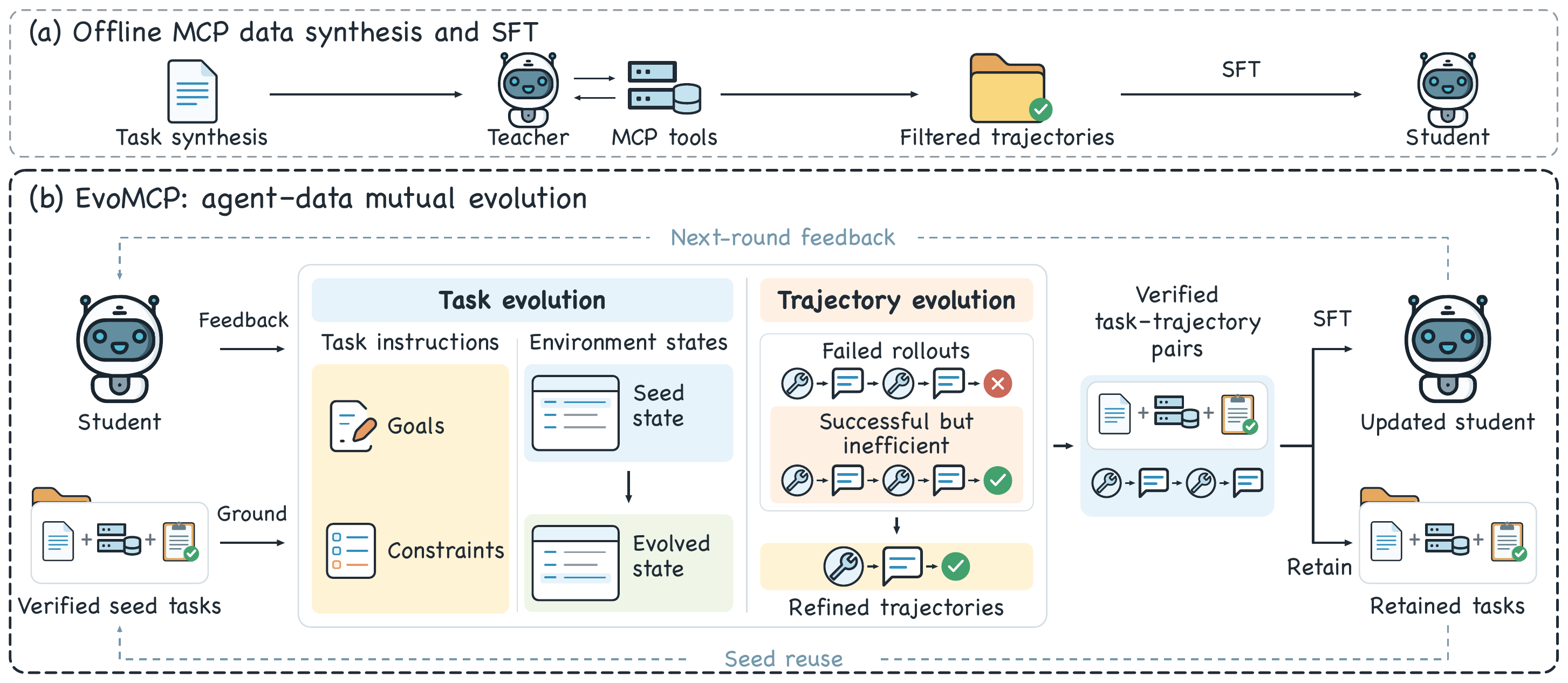}
\caption{\textbf{Self-evolving training in real MCP environments.} (a) TOUCAN synthesizes tasks and filters teacher trajectories executed through real MCP tools for SFT \citep{xu2025toucansynthesizing15mtoolagentic}. (b) \method\ uses student rollout feedback to assess capabilities and select learning targets and verified seed tasks. Task evolution adapts task instructions, environment states, or both while keeping verifiers aligned. Trajectory evolution uses reusable skills to recover failed teacher rollouts and refine successful demonstrations. Accepted demonstrations fine-tune the student, and accepted tasks supply seeds for later rounds.}
\label{fig:overview}
\end{figure}

As shown in Figure~\ref{fig:overview}, \method\ couples task evolution with trajectory evolution to adapt both training tasks and teacher demonstrations. Within task evolution, instruction evolution revises task goals or strengthens constraints. Environment evolution provides a complementary transformation by varying observable execution states, such as data values and object relations, while preserving the capability goal and MCP tool interfaces. These transformations can be used separately or together, with instructions, environments, and verifiers kept mutually consistent. To provide reliable and efficient supervision for these tasks, trajectory evolution uses reusable skills to recover failed teacher rollouts and refine successful ones. Together, these steps allow us to transform seed tasks and teacher rollouts into high-quality training data, enabling more reliable supervised fine-tuning of MCP agents.

\looseness=-1
After validating the resulting task--trajectory pairs, we use accepted demonstrations to fine-tune the student, and the corresponding tasks become seed tasks for subsequent evolution. The updated student is then re-evaluated to provide fresh feedback for selecting learning targets and compatible seeds in the next round. This closes the loop between data evolution and student learning, allowing the training distribution to adapt as the student's capabilities change. We evaluate \method\ on three representative benchmarks, Toolathlon \citep{li2026tooldecathlonbenchmarkinglanguage}, MCPMark \citep{ICLR2026_8138d211}, and MCP-Atlas \citep{bandi2026mcpatlaslargescalebenchmarktooluse}, using Qwen3.6-27B as the backbone. By continuously synthesizing new tasks from training-time feedback and refining teacher demonstrations, \method\ achieves absolute gains of 16.5\%, 10.2\%, and 12.6\% over the initial student, respectively, demonstrating the effectiveness of agent--data mutual evolution across MCP environments. Our contributions can be summarized as follows:
\begin{itemize}
  \item We propose \method, an agent--data mutual evolution framework that alternates between student fine-tuning and the evolution of complete MCP task--trajectory pairs, reusing verified tasks as seeds across rounds.
  \item Unlike static data synthesis, we incorporate student rollout feedback into learning-target selection and task evolution from verified seeds, and use reusable skills to recover failed teacher rollouts and refine successful demonstrations.
  \item \method\ yields consistent gains over the initial student and static verified SFT across three MCP benchmarks, demonstrating its effectiveness in complex tool-use environments.
\end{itemize}

\section{Related Work}
\label{sec:related}

\paragraph{Task and environment synthesis.}
MCP data pipelines synthesize tasks and collect tool-use trajectories through real MCP servers \citep{xu2025toucansynthesizing15mtoolagentic,wang-etal-2026-mcp}.
For task construction, AgentGen \citep{hu2025agentgenenhancingplanningabilities} grounds generation in environment descriptions, while TaskCraft \citep{shi2025taskcraftautomatedgenerationagentic} composes simpler tasks into complex workflows.
EnvScaler \citep{song2026envscalerscalingtoolinteractiveenvironments} and CLI-Universe \citep{hua2026cliuniverseverifiabletasksynthesis} jointly synthesize executable environments, tasks, and validators.
RST \citep{li2026recursivesynthesislonghorizonterminal} realigns the components of terminal tasks and reuses validated tasks as seeds across synthesis generations.
Our work builds on this line by evolving complete MCP tasks from verified seeds through complementary changes to task instructions and observable environment states, keeping their verifiers aligned.

\paragraph{Feedback-driven agent training.}
Feedback-driven methods adapt the training distribution as the agent changes.
Tool-R0 \citep{acikgoz2026toolr0selfevolvingllmagents} and GenEnv \citep{guo2025genenvdifficultyalignedcoevolutionllm} target learner-relative difficulty; LoopTool \citep{zhang-etal-2026-looptool} generates data around diagnosed tool-call errors, while CoEvolve \citep{yang-etal-2026-coevolve} uses forgetting and uncertainty from rollouts to guide task synthesis.
Agent-World \citep{dong2026agentworldscalingrealworldenvironment} uses capability diagnosis to guide task and environment expansion.
In \method, student rollout feedback selects learning targets and compatible verified seeds to determine which tasks to reuse and which new tasks to construct.

\paragraph{Trajectory synthesis and refinement.}
Teacher failures limit demonstration coverage, while errors in successful rollouts affect supervision quality.
Trial and Error \citep{song-etal-2024-trial} forms preference pairs from successful and failed attempts, while STeCa \citep{wang-etal-2025-steca} calibrates suboptimal actions through step-level feedback.
For successful trajectories, WebSTAR \citep{he-etal-2026-webstar} masks low-quality steps in the SFT loss.
Complementary approaches retain reflections, insights, or skills to guide later attempts, as in Reflexion \citep{shinn2023reflexionlanguageagentsverbal}, ExpeL \citep{zhao2024expelllmagentsexperiential}, and Ace-Skill \citep{xiong2026aceskillbootstrappingmultimodalagents}.
\method\ uses reusable skills to recover failed teacher rollouts on fixed tasks and refine successful demonstrations to reduce tool errors and avoidable work.
Verified demonstrations then fine-tune the student, while accepted tasks become seeds for later evolution.
Updated student feedback guides the next round.

\section{Problem Formulation}
\label{sec:problem}

\paragraph{MCP tasks and training data.}
An MCP task $x=(q,E,V)$ consists of a task instruction $q$ specifying the goal and constraints, an execution environment $E$ with an initial state and available MCP tools, and a task verifier $V$. To solve the task, an agent interacts with $E$ according to $q$, producing a trajectory $\tau$ of messages, tool calls, and observations. The verifier returns $V(\tau)=1$ if the execution satisfies the task's success criteria and $0$ otherwise. For training, we use accepted task--trajectory pairs $d=(x,\tau)$, where $x$ passes the applicable task checks and $\tau$ is a teacher demonstration satisfying $V(\tau)=1$.

\paragraph{Student feedback.}
\method\ alternates data construction and student training over evolution rounds. At the start of round $t$, the student policy $\pi_t$ performs rollouts on a set of MCP tasks. The resulting trajectories and task outcomes yield feedback signals $\mathcal{S}_t$ that characterize the student's established capabilities and capability gaps. This assessment identifies behaviors to reinforce or improve, which become the learning targets for data construction. We then select verified seed tasks whose requirements, tools, and resources support these targets.

\paragraph{Agent--data mutual evolution.}
At evolution round $t$, we maintain an accumulated dataset $\D_t$ of verified task--trajectory pairs. The initial dataset $\D_0$ contains verified seed tasks and their teacher demonstrations. Using the learning targets and seed tasks selected above, \method\ constructs the training set $\D_t^{\mathrm{train}}$ through task and trajectory evolution. We complete the round by using the selected demonstrations for SFT and adding newly accepted pairs to $\D_t$, giving
\begin{equation}
\begin{aligned}
\pi_{t+1}&=\operatorname{SFT}(\pi_t,\D_t^{\mathrm{train}}),\\
\D_{t+1}&=\D_t\cup\D_t^{\mathrm{train}}.
\end{aligned}
\label{eq:selfevolution}
\end{equation}
Re-evaluating the updated student $\pi_{t+1}$ yields fresh feedback for selecting learning targets in the next round. The expanded dataset $\D_{t+1}$ supplies verified seed tasks for these targets, so subsequent data construction is guided by both the updated student and the accumulated data.

\section{Methodology}
\label{sec:method}

\begin{figure}[t]
\centering
\includegraphics[width=\textwidth]{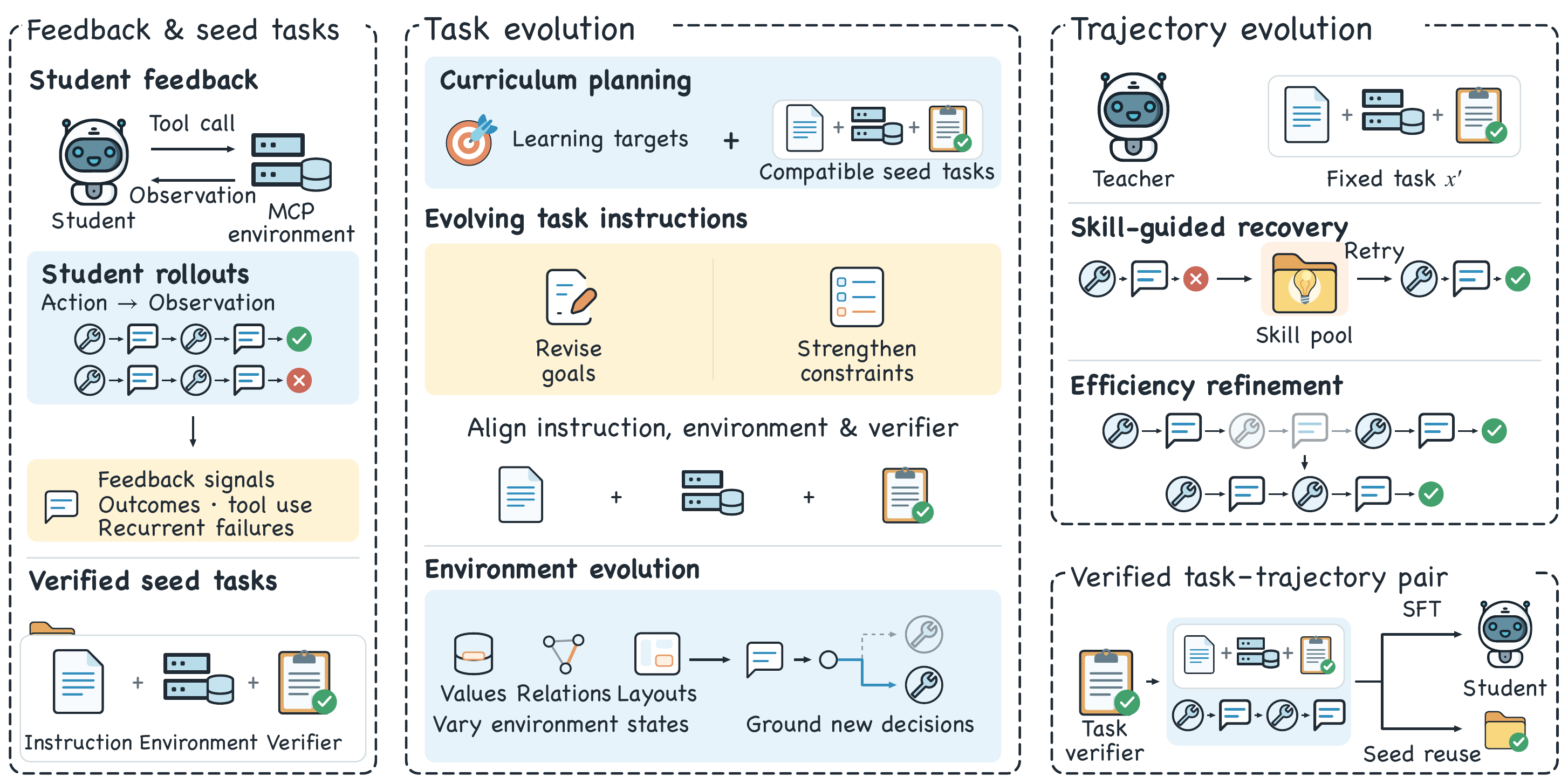}
\caption{\textbf{Overview of \method.} A curriculum planner uses student feedback to select learning targets and verified seed tasks. Task evolution adapts task instructions, environment states, or both while keeping their verifiers aligned. Trajectory evolution uses reusable skills to recover failed teacher rollouts and refine successful demonstrations on each fixed task. Accepted demonstrations train the student, and the corresponding tasks become seed tasks for subsequent rounds.}
\label{fig:experience-verification}
\end{figure}

As shown in Figure~\ref{fig:experience-verification}, \method\ couples student-guided task evolution with trajectory evolution to adapt both training tasks and teacher demonstrations. Our core idea is to evolve complete task--trajectory pairs by adapting verified seed tasks to learning targets identified from student rollout feedback and improving the teacher demonstrations used for SFT. Task evolution adapts task instructions, environment states, or both while keeping verifiers aligned. We collect teacher rollouts on these evolved tasks to provide demonstrations for SFT. Trajectory evolution then uses reusable skills to recover failed teacher rollouts and refine successful demonstrations. Accepted demonstrations fine-tune the student, while the corresponding tasks seed later rounds.

\subsection{Student-Guided Task Evolution}
\label{sec:task-evolution}

We assess the student using verifier outcomes, tool-use patterns, and recurrent failures from its rollouts \citep{yang-etal-2026-coevolve}. The resulting profile identifies established behaviors and capability gaps, linking observed weaknesses to failure mechanisms and supporting rollouts. A curriculum planner uses this profile to select learning targets for retaining learned behaviors, correcting weaknesses, or exploring behaviors with limited evidence. It matches these targets to verified seed tasks whose requirements, tools, and resources support the intended behavior, together with their verified demonstrations from $\D_t$. The planner may reuse an existing pair or evolve its task. For evolution, it selects the direction of change, while a task generator determines the concrete requirements and state changes supported by the selected seed tasks.

When evolving the task from a single verified seed pair $d=(x,\tau)\in\D_t$, the learning target derived from feedback $s\in\mathcal{S}_t$ guides two complementary transformations of $x=(q,E,V)$.
\begin{equation}
\begin{aligned}
\mathcal{T}_{s}(x)&=(q',E_{\!T},V_{\!T}),\\
\mathcal{E}_{m}(x)&=(q_m,E_m,V_m).
\end{aligned}
\label{eq:experience-construction}
\end{equation}
For a selected learning target, $\mathcal{T}_{s}$ changes task requirements, while $\mathcal{E}_{m}$ produces task instances with varied observable states. Here, $m$ specifies the state elements to vary, such as data values or object relations. The planner can select either route or apply a compatible environment transformation to the task produced by $\mathcal{T}_{s}$. We denote the resulting task by $x'$.

\paragraph{Evolving task instructions.}
Conditioned on the selected learning target and compatible verified seed tasks, we prompt the task generator to specify the behavior to introduce and the requirements to retain. Specifically, it revises task goals or strengthens constraints based on the tools and resources of the selected seed tasks. For example, a goal revision can combine subgoals and introduce dependencies between them. The framework then generates $q'$ and updates resources in $E_{\!T}$ as needed, binding task references to information accessible through tool interactions.

Accordingly, we retain $V_{\!T}$ when the success criteria are unchanged, and extend or revise its checks when requirements are added or the objective changes. We then check consistency among $q'$, $E_{\!T}$, and $V_{\!T}$. Together with evidence from the verified seed tasks, resource checks and positive and negative verifier tests support solvability assessment before teacher collection. This process yields tasks tailored to the selected learning targets, with aligned instructions, environments, and verifiers.

\paragraph{Environment evolution.}
\label{sec:environment-evolution}

In addition to changing task requirements, we evolve environment states to provide diverse task instances for the selected learning targets. The motivation is that the same capability can require different information, tool arguments, or decisions under different observable conditions. For example, we vary task-relevant data values, object relations, or layouts.

To keep each instance consistent, we update references in $q_m$ and the factual and resource bindings of $V_m$ as needed to match $E_m$, while preserving the parent task's success criteria. We collect teacher demonstrations for the selected learning targets across these varied observable states.

\subsection{Trajectory Evolution}
\label{sec:trajectory-evolution}

Task validity alone does not ensure a useful demonstration: teacher rollouts may fail, while successful ones can contain suboptimal intermediate actions that full-trajectory SFT directly supervises \citep{he-etal-2026-webstar}. We recover failed rollouts and refine successful ones while keeping task $x'$ fixed.

\paragraph{Skill-guided failure recovery.}
We first distinguish model-behavior failures from task, verifier, and runtime failures. Only model-behavior failures enter skill-guided recovery. For such failures, \method{} routes a compatible active skill. If none applies, it distills a new recovery skill from the failed rollout and agent-visible feedback. Each skill encodes an observable failure condition, a corrective procedure grounded in agent-visible information, and a completion check. The teacher retries in a fresh session, and only verifier-passing trajectories are retained. Newly distilled skills enter the pool only after passing the unchanged verifier on their source task.

\paragraph{Efficiency refinement.}
Successful trajectories are audited for real tool errors and avoidable work. Let $N_{\mathrm{red}}(\tau)$ and $N_{\mathrm{succ}}(\tau)$ denote redundant and all successful tool calls. For assistant segment $j$, let $a_j$ be its avoidable fraction and $n_j$ its length. We define the \emph{avoidable-work ratio} (AWR) as
\begin{equation}
\operatorname{AWR}(\tau)
=
\max\!\left\{
\frac{N_{\mathrm{red}}(\tau)}{\max\{N_{\mathrm{succ}}(\tau),1\}},
\frac{\sum_j n_j a_j}{\max\{\sum_j n_j,1\}}
\right\}.
\label{eq:avoidable-work-ratio}
\end{equation}
If a trajectory misses its task-specific tool-error or AWR threshold, the teacher reruns the unchanged task using the audit findings as teacher-only guidance. Among verifier-passing candidates, we prioritize fewer tool errors and then lower AWR.

\subsection{Validation, Training, and Reuse}
\label{sec:export}

Before SFT, we exclude teacher-only skill guidance and judge annotations from the training examples so that the training context contains the task inputs available to the student. For an evolved task $x'$, the selected and cleaned trajectory $\tau^\star$ forms a candidate pair $d'=(x',\tau^\star)$. It enters $\D_t^{\mathrm{train}}$ only if the task passes validation checks and the teacher execution has passed the task verifier. We also check that the exported messages preserve a consistent sequence of tool calls and observations. Verified task--trajectory pairs selected for direct reuse enter $\D_t^{\mathrm{train}}$ with their existing demonstrations.

Following Equation~\ref{eq:selfevolution}, demonstrations in $\D_t^{\mathrm{train}}$ supervise the update from $\pi_t$ to $\pi_{t+1}$, and newly accepted task--trajectory pairs expand $\D_{t+1}$. Re-evaluating the updated student through the same execution framework yields fresh feedback that, together with the expanded seed pool, guides task selection and evolution in the next round.

\section{Experiments}
\label{sec:experiments}

\subsection{Experimental Setup}

\paragraph{Benchmarks and metrics.}
We evaluate on Toolathlon \citep{li2026tooldecathlonbenchmarkinglanguage}, MCPMark \citep{ICLR2026_8138d211}, and MCP-Atlas \citep{bandi2026mcpatlaslargescalebenchmarktooluse}. They cover long-horizon, stateful tool use, cross-server orchestration, and claim-level synthesis. We report Toolathlon task success and MCPMark and MCP-Atlas Pass@1, using the official 0.75 claim-coverage threshold for the latter.

\paragraph{Data and evaluation.}
Verified seed task--trajectory pairs initialize $\mathcal{D}_0$. Benchmark rollouts provide student feedback for learning-target and seed-task selection, and successive checkpoints are evaluated on the same benchmark task sets. We refresh this feedback after each student update. All student checkpoints use the same evaluation setup. For MCP-Atlas, we use GLM-5.2 as the claim-level judge, with its prompt and decoding settings held fixed.

\paragraph{Models and training.}
We use Qwen3.6-27B as the student and Qwen3.8-Max as the task generator and teacher. R0 denotes the initial student, and R1--R3 denote the checkpoints after one to three rounds of full-parameter SFT. We train for three epochs on 2,400 verifier-accepted demonstrations per round. We use a learning rate of $2\times10^{-6}$ and a global batch size of 128. Further implementation details and results appear in the appendix.

\subsection{Main Results}

\begin{table}[!t]
\caption{\textbf{Performance on three tool-use benchmarks (\%).} Scores are task success for Toolathlon and Pass@1 for MCPMark and MCP-Atlas. Bold and underlined values mark the highest and second-highest reported scores per benchmark, respectively. Blue rows denote \method{} checkpoints. Params gives total/activated parameters. Public reference scores use their original evaluation settings.}
\label{tab:model-comparison}
\centering
\begin{tabularx}{\textwidth}{@{}Ycccc@{}}
\toprule
\textbf{Model} & \textbf{Params} & \textbf{Toolathlon} & \textbf{MCPMark} & \textbf{MCP-Atlas} \\
\midrule
\multicolumn{5}{c}{\textit{Closed-source agents}} \\
\midrule
GPT-5.4 (xhigh) & -- & \textbf{54.6} & 62.5 & 70.6 \\
Claude Opus 4.5 & -- & 43.5 & 42.6 & 69.8 \\
Gemini 3 Pro & -- & 36.4 & 53.9 & 70.3 \\

\midrule
\multicolumn{5}{c}{\textit{Open-weight models}} \\
\midrule
Kimi K2.5 & 1T/32B & 27.8 & 29.5 & 64.4 \\
DeepSeek-V4-Flash & 284B/13B & 47.8 & 63.8 & 69.0 \\
\midrule
\multicolumn{5}{c}{\textit{Student checkpoints}} \\
\midrule
Qwen3.6-27B & 27B/27B & 36.9 & 56.7 & 61.0 \\
\rowcolor{oursrow}
\quad + \method (R1) & 27B/27B & 46.6 & 60.6 & 68.4 \\
\rowcolor{oursrow}
\quad + \method (R2) & 27B/27B & 50.5 & \underline{64.5} & \underline{71.3} \\
\rowcolor{oursrow}
\quad + \method (R3) & 27B/27B & \underline{53.4} & \textbf{66.9} & \textbf{73.6} \\
\bottomrule
\end{tabularx}
\end{table}

Table~\ref{tab:model-comparison} compares public reference models with the initial student and successive \method{} checkpoints. Public-model scores and parameter counts follow official reports and model cards \citep{deepseekai2026deepseekv4highlyefficientmilliontoken}, using their source-specific harnesses and inference budgets.

\paragraph{Overall performance.}
\method{} consistently improves student performance across all three benchmarks. After three rounds, it reaches 53.4\% on Toolathlon, 66.9\% on MCPMark, and 73.6\% on MCP-Atlas, yielding gains of 16.5\%, 10.2\%, and 12.6\% over R0. Performance improves in every round on all three benchmarks, showing that successive rounds of data evolution continue to benefit the updated student. We next compare against continued training on fixed seed data to examine the contribution of evolving the training pairs.

\subsection{Ablation Study}

\paragraph{Comparison settings.}
We compare \method{} with \emph{Static SFT}. Starting from R0, this baseline repeatedly trains on the same verifier-passing seed-task demonstrations. At R1, R2, and R3, its cumulative training steps match those of the corresponding \method{} checkpoint.

At R1, we further remove one component at a time. \emph{Without student feedback}, learning-target selection, seed selection, transformation planning, and task construction exclude student outcomes and capability diagnoses, relying on seed semantics, available resources, and task diversity. \emph{Without task evolution}, the selected verified tasks are reused unchanged. \emph{Without trajectory evolution}, we retain the first verifier-passing teacher rollout without recovery or refinement. All variants use the same initial checkpoint, number of demonstrations, and SFT settings. The remaining components follow the full method.

\begin{table}[!t]
\caption{\textbf{Component ablations at R1 and Static SFT comparisons through R3.} Cumulative training steps match per round. Scores (\%) use the metrics in Table~\ref{tab:model-comparison}. Checkmarks indicate enabled components: student feedback (Stud.~fb.), task evolution (Task~evo.), and trajectory evolution (Traj.~evo.). Bold marks each round's best scores, and shading denotes \method{}.}
\label{tab:training-comparison}
\centering
\small
\setlength{\tabcolsep}{2.2pt}
\renewcommand{\arraystretch}{1.06}
\begin{tabularx}{\textwidth}{@{}>{\normalsize\raggedright\arraybackslash}p{1.44in}>{\centering\arraybackslash}p{0.54in}>{\centering\arraybackslash}p{0.49in}>{\centering\arraybackslash}p{0.54in}*{3}{>{\centering\arraybackslash}X}@{}}
\toprule
\textbf{Variant} & \textbf{Stud. fb.} & \textbf{Task evo.} & \textbf{Traj. evo.} & \textbf{Toolathlon} & \textbf{MCPMark} & \textbf{MCP-Atlas} \\
\midrule
Initial student (R0) & -- & -- & -- & 36.9 & 56.7 & 61.0 \\
\midrule
\multicolumn{7}{l}{\textit{Round 1}} \\
Static SFT & $\times$ & $\times$ & $\times$ & 40.8 & 57.4 & 65.8 \\
w/o student feedback & $\times$ & $\checkmark$ & $\checkmark$ & 42.2 & 58.2 & 66.4 \\
w/o task evolution & $\checkmark$ & $\times$ & $\checkmark$ & 42.4 & 57.4 & 66.2 \\
w/o trajectory evolution & $\checkmark$ & $\checkmark$ & $\times$ & 44.3 & 59.5 & 67.8 \\
\rowcolor{oursrow}
\textbf{\method{}} & $\boldsymbol{\checkmark}$ & $\boldsymbol{\checkmark}$ & $\boldsymbol{\checkmark}$ & \textbf{46.6} & \textbf{60.6} & \textbf{68.4} \\
\midrule
\multicolumn{7}{l}{\textit{Round 2}} \\
Static SFT & $\times$ & $\times$ & $\times$ & 40.8 & 57.4 & 67.2 \\
\rowcolor{oursrow}
\textbf{\method{}} & $\boldsymbol{\checkmark}$ & $\boldsymbol{\checkmark}$ & $\boldsymbol{\checkmark}$ & \textbf{50.5} & \textbf{64.5} & \textbf{71.3} \\
\midrule
\multicolumn{7}{l}{\textit{Round 3}} \\
Static SFT & $\times$ & $\times$ & $\times$ & 39.8 & 58.2 & 67.2 \\
\rowcolor{oursrow}
\textbf{\method{}} & $\boldsymbol{\checkmark}$ & $\boldsymbol{\checkmark}$ & $\boldsymbol{\checkmark}$ & \textbf{53.4} & \textbf{66.9} & \textbf{73.6} \\
\bottomrule
\end{tabularx}
\end{table}

\paragraph{Effect of iterative evolution.}
Table~\ref{tab:training-comparison} shows that \method{} outperforms Static SFT at every round. Its absolute gains grow from 5.8\%, 3.2\%, and 2.6\% at R1 to 13.6\%, 8.7\%, and 6.4\% at R3 on Toolathlon, MCPMark, and MCP-Atlas, respectively. With cumulative training steps matched, these results support refreshing the task--trajectory pairs as the student improves, extending the benefit of data evolution beyond the first round.

\paragraph{Contribution of individual components.}
At R1, removing any component reduces scores on all three benchmarks. The larger drops follow the removal of student feedback (2.0--4.4 points) or task evolution (2.2--4.2 points), highlighting the value of selecting learning targets from student behavior and constructing tasks to realize them. Removing trajectory evolution also lowers scores by 0.6--2.3 points, showing a further benefit from recovering and refining teacher demonstrations. Together, these results support combining feedback-guided task construction with trajectory evolution.

\subsection{Analysis of Task and Trajectory Evolution}

\begin{table}[!t]
\caption{\textbf{Effects of task and trajectory evolution.} (a) Seed--evolved task pairs, with $\Delta$ denoting evolved-minus-seed differences in percentage points. (b) Matched verifier-passing teacher demonstrations before and after refinement. AWR denotes avoidable-work ratio. Down arrows indicate lower is better. Shading and boldface identify the evolved condition.}
\label{tab:mechanisms}
\centering
\footnotesize
\setlength{\tabcolsep}{5pt}
\renewcommand{\arraystretch}{1.12}
\begin{tabularx}{\textwidth}{@{}Y>{\centering\arraybackslash}p{1.02in}>{\centering\arraybackslash}p{1.08in}>{\centering\arraybackslash}p{0.72in}@{}}
\toprule
\multicolumn{4}{c}{\textbf{(a) Task evolution}} \\
\midrule
\textbf{Metric} & \textbf{Seed} & \cellcolor{oursrow}\textbf{Evolved} & \textbf{$\Delta$} \\
\midrule
Target realization (\%) & 27.8 & \cellcolor{oursrow}\textbf{85.7} & +57.9 \\
Behavior-changing tasks (\%) & -- & \cellcolor{oursrow}\textbf{100.0} & -- \\
Student failure rate (\%) & 24.5 & \cellcolor{oursrow}\textbf{45.3} & +20.8 \\
Teacher success rate (\%) & 89.7 & \cellcolor{oursrow}\textbf{78.9} & $-10.8$ \\
\bottomrule
\end{tabularx}

\vspace{0.7em}

\begin{tabularx}{\textwidth}{@{}Y>{\centering\arraybackslash}p{1.25in}>{\centering\arraybackslash}p{1.25in}@{}}
\toprule
\multicolumn{3}{c}{\textbf{(b) Trajectory evolution}} \\
\midrule
\textbf{Metric} & \textbf{w/o Traj. Evo.} & \cellcolor{oursrow}\textbf{w/ Traj. Evo.} \\
\midrule
Mean tool-call errors $\downarrow$ & 2.6 & \cellcolor{oursrow}\textbf{1.6} \\
Mean AWR (\%) $\downarrow$ & 31.1 & \cellcolor{oursrow}\textbf{30.6} \\
Mean output tokens $\downarrow$ & 9,752.0 & \cellcolor{oursrow}\textbf{8,873.0} \\
\bottomrule
\end{tabularx}
\end{table}

\paragraph{Effect of task evolution.}
Table~\ref{tab:mechanisms}(a) compares seed and evolved tasks at R1 against the same feedback-derived learning target for each pair. Target realization checks whether the intended behavior is required by the task, supported by the environment, and tested by the verifier. Behavior change measures whether evolution alters the required actions or decisions. We also evaluate student failure and teacher success using R0 and the same teacher under matched execution budgets, before recovery or refinement.

Task evolution increases target realization from 27.8\% to 85.7\%, and all sampled evolved tasks require changed solution behavior. It also raises R0's failure rate from 24.5\% to 45.3\%, while the teacher still succeeds on 78.9\% of evolved tasks, compared with 89.7\% of seed tasks. Together, these results show that evolved tasks more consistently instantiate the selected learning targets, expose unmet demands for the student, and remain solvable by the teacher in most cases. This connects student feedback to executable tasks for collecting successful teacher demonstrations.

For example, an evolved analyst-rating task requires pooling observations across three forecast horizons before computing a hit rate, where the seed reports each horizon separately. Its updated verifier checks the pooled statistics, observation counts, and report structure, and the collected teacher demonstration passes. This case illustrates how a numeric-accuracy target changes the required aggregation while reusing the source data.

\paragraph{Effect of trajectory evolution.}
Table~\ref{tab:mechanisms}(b) evaluates trajectory refinement at R1 on matched verifier-passing teacher demonstrations. Refinement reduces mean tool-call errors from 2.6 to 1.6 and mean output tokens from 9,752.0 to 8,873.0, a 9.0\% reduction. Mean AWR changes more modestly, from 31.1\% to 30.6\%. These results show that refinement yields more concise successful demonstrations with fewer tool-call errors. The gains in Table~\ref{tab:training-comparison} further support the combined trajectory-evolution stage for student SFT.

\begin{figure}[!t]
\centering
\includegraphics[width=\textwidth,trim=0 18bp 0 0,clip]{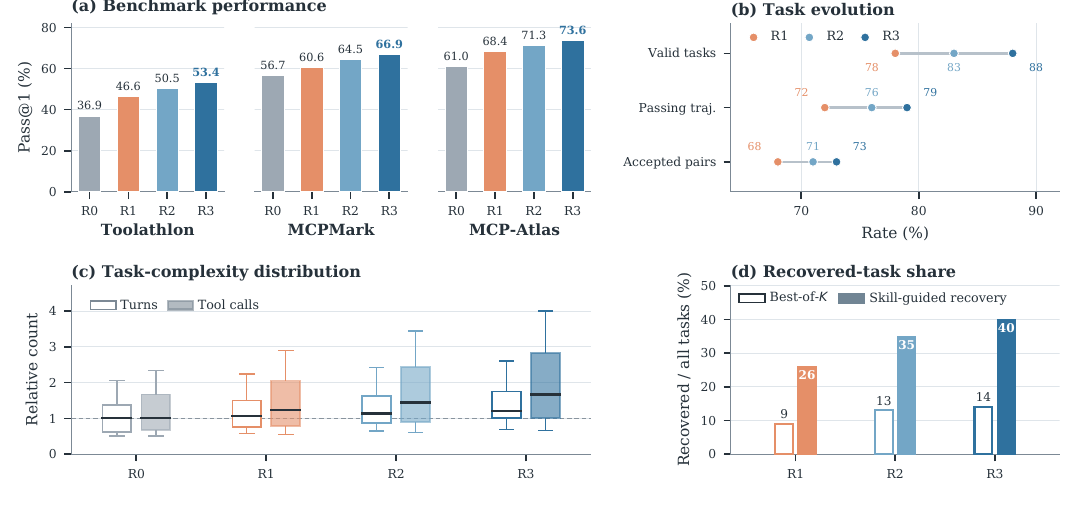}
\caption{\textbf{Evolution dynamics across rounds.} (a) Benchmark performance from R0 to R3. (b) Task-validity, trajectory-acceptance, and training-pair-admission rates. (c) Turns and tool calls, each normalized by its R0 median, as proxies for interaction workload. (d) Recovered tasks as a percentage of all tasks in each round under a matched teacher-rollout budget. Best-of-K denotes unguided retries.}
\label{fig:evolution-dynamics}
\end{figure}

\paragraph{Evolution dynamics.}

Figure~\ref{fig:evolution-dynamics} tracks student performance and training-data changes across rounds. In panel (b), task validity increases from 78.0\% at R1 to 88.0\% at R3, trajectory acceptance from 72.0\% to 79.0\%, and training-pair admission from 68.0\% to 73.0\%. The improvement extends from task construction through teacher collection to the admission of verified training pairs. Thus, later rounds yield a higher proportion of usable supervision alongside the continued student gains in panel (a). These trends complement the gains over Static SFT in Table~\ref{tab:training-comparison}, where cumulative training steps are matched at each checkpoint.

Panel (c) shows how the teacher's interaction workload changes. Both the median and upper quartile of tool-call counts increase, while turns grow more moderately. The distribution therefore shifts toward executions with more tool interaction, extending the change beyond a few unusually long trajectories. Together with panel (b), this shows that the pipeline continues to produce usable training pairs as teacher executions become longer and involve more tool use.

Panel (d) compares skill-guided recovery with unguided retries under a matched teacher-rollout budget. The recovered-task share grows from 26.0\% at R1 to 40.0\% at R3 with skill guidance, compared with 9.0\% to 14.0\% for unguided retries. The absolute advantage therefore widens from 17.0\% to 26.0\% across rounds. This advantage is present from the first round and persists as the task pool evolves, supporting reusable recovery guidance for obtaining demonstrations from initially unsuccessful executions. Alongside Table~\ref{tab:mechanisms}(b), these results highlight the two roles of trajectory evolution: recovering successful demonstrations for initially failed tasks and refining the efficiency of already successful demonstrations. Together, task adaptation and teacher-demonstration improvement produce verified experience for both student training and subsequent task construction, linking the gains within each round to continued data evolution across rounds.

\section{Conclusion}

We introduce \method, a framework for co-evolving MCP agents and verified task--trajectory pairs through supervised fine-tuning. By using student feedback to guide task construction and improving teacher demonstrations through recovery and refinement, \method{} adapts its training data across rounds. Accepted demonstrations train the student, while accepted tasks seed subsequent evolution. Experiments on three tool-use benchmarks show consistent gains across three rounds and a widening advantage over Static SFT under matched cumulative training steps. Component ablations further support the roles of student feedback, task evolution, and trajectory evolution. Together, these findings highlight task construction and teacher supervision as complementary components of iterative MCP-agent training, with verified experience connecting data evolution and student learning.

\label{page:end-main}
\clearpage
\bibliography{references}
\bibliographystyle{iclr2027_conference}
\clearpage
\appendix
\addtocontents{toc}{\protect\setcounter{tocdepth}{2}}
\raggedbottom
\renewcommand{\contentsname}{Appendix Contents}
\begingroup
\fontsize{11}{13}\selectfont
\color{promptblue}
\hypersetup{pdfborder={0 0 0}}
\setlength{\parskip}{0pt}
\makeatletter
\renewcommand{\l@section}[2]{\addpenalty{\@secpenalty}\vskip 4pt
  \@dottedtocline{1}{0em}{2.2em}{\bfseries #1}{\textcolor{promptblue}{#2}}}
\renewcommand{\l@subsection}[2]{\@dottedtocline{2}{2.2em}{3.0em}{#1}{\textcolor{promptblue}{#2}}}
\makeatother
\tableofcontents
\endgroup
\clearpage

\section{Training and Implementation Details}
\label{app:training_details}
\label{app:verification}

\subsection{Training configuration}
\label{app:training_configuration}

Table~\ref{tab:training-configuration} reports the SFT settings used in each
evolution round. Static SFT repeatedly uses the same seed demonstrations,
matching the cumulative training steps at each checkpoint.

\begin{table}[H]
\caption{Supervised fine-tuning configuration.}
\label{tab:training-configuration}
\centering
\footnotesize
\setlength{\tabcolsep}{7pt}
\renewcommand{\arraystretch}{0.96}
\begin{tabularx}{0.82\linewidth}{@{}Y>{\centering\arraybackslash}X@{}}
\toprule
\textbf{Setting} & \textbf{Qwen3.6-27B} \\
\midrule
Training method & Full-parameter SFT \\
Training trajectories per round & 2,400 \\
Epochs per round & 3 \\
Optimizer & Adam \\
Adam betas & $(0.9,0.95)$ \\
Weight decay & $0.1$ \\
Learning rate & $2\times10^{-6}$ \\
Minimum learning rate & $1\times10^{-8}$ \\
Schedule & Cosine decay \\
Warmup ratio & 10\% \\
Micro batch size & 1 \\
Global batch size & 128 \\
Maximum sequence length & 131,072 \\
Hardware & 128 PPU-810E \\
Tensor parallelism & 4 \\
Context parallelism & 2 \\
Data parallelism & 16 \\
\bottomrule
\end{tabularx}
\end{table}

\subsection{MCP-Atlas evaluation protocol}
\label{app:atlas_judge}

All MCP-Atlas results for our student checkpoints use GLM-5.2 as the
claim-level judge under one fixed scoring protocol. Each reference claim is
evaluated independently against the student's final answer and assigned one
of three outcomes: fulfilled, partially fulfilled, or not fulfilled, mapped to
$1$, $0.5$, and $0$, respectively. Claim coverage is the mean score across all
reference claims for a task, and the task passes when coverage is at least
$0.75$.

Judge calls use temperature $0$, a maximum output length of 4,096 tokens, and
a strict JSON schema containing the claim, outcome, justification, and
confidence. Empty or failed student responses receive zero coverage. The judge model, prompt, decoding settings, outcome mapping, and pass threshold are held fixed across all student checkpoints.

\subsection{Task construction and validation}
\label{app:task_implementation}

The curriculum planner receives the student capability report and the available
seed roster. It allocates retain, remediate, and explore lanes, selects compatible
runtime seeds, and specifies a mutation axis and change strength. Dispatch turns
these choices into individual work orders. The planner selects the learning
direction; the task builder determines the concrete requirement change from the
selected seeds and their executable materials.

The builder receives the work order, the complete seed request, indexed source
and implementation materials, available operations and MCP tools, and previously
generated sibling goals. It can inspect missing source facts before returning an
operation with its parameters. For a local change, the implementation preserves
the seed setup and unaffected evaluator checks, and patches the changed rule and
its tests. Typed task programs additionally encode goals, dependencies,
constraints, relevant tools, and preserved clauses; their surface realizer
produces one coherent request. Public requests state the required outcome without
revealing student diagnoses, private answers, or evaluator internals.

Environment evolution operates on a copied initial state. Changed records and
resources are bound back to the request, while the MCP interface and unchanged
success criteria are preserved. When a state change alters an observable
requirement, the corresponding verifier input is updated with the same binding.
This implementation treats the request, state, and verifier as one candidate;
none is admitted independently.

Each candidate retains its source seeds, capability-report reference, planner
assignment, construction record, and execution receipts. Environment variants
used for held-out evaluation are kept outside the curriculum and training export.
The cases in Section~\ref{app:task_evolution_cases} illustrate both an environment
pair from this evaluation split and a requirement change applied to an evolved
training seed.

Validation checks that the initial state materializes, declared tools are
available, and task references resolve to observable resources. Positive tests
must satisfy the intended requirements, while focused counterexamples that omit
or corrupt a changed requirement must fail. Before teacher collection, we assess
solvability using evidence from the verified seeds and these positive and negative
tests. Inclusion in the training set additionally requires a teacher demonstration
that passes the candidate's native verifier.

\subsection{Failure attribution and skill routing}
\label{app:trajectory_evolution}
\label{app:failure_attribution}
\label{app:skill_construction}
\label{app:skill_routing}

A failed teacher rollout is assigned to model behavior, the task or verifier
contract, the execution environment, the harness, or an unresolved mixture.
Only model-behavior failures enter skill-guided recovery. The diagnosis records
the first problematic step, failed requirements, supporting evidence, failure
type and subtype, a minimal correction, and confidence. Skill construction
requires non-empty evidence and confidence of at least $0.8$.

Table~\ref{tab:trajectory_failure_taxonomy} lists the behavioral taxonomy used
for diagnosis and routing.

\begin{table}[H]
\centering
\footnotesize
\setlength{\tabcolsep}{4pt}
\renewcommand{\arraystretch}{1.08}
\caption{Trajectory-failure taxonomy used for diagnosis and skill routing.}
\label{tab:trajectory_failure_taxonomy}
\begin{tabularx}{\linewidth}{@{}L{0.18\linewidth}L{0.34\linewidth}Y@{}}
\toprule
\textbf{Type} & \textbf{Representative subtypes} & \textbf{Boundary} \\
\midrule
Planning & Turn-budget exhaustion; redundant search; incorrect ordering
& Available actions are arranged into an ineffective plan. \\
Tool selection & Wrong tool; omitted required tool
& An exposed operation is omitted or replaced by an unsuitable one. \\
Argument & Missing argument; wrong scope; invalid schema
& The selected tool is appropriate, but its arguments are not. \\
Recovery & No retry; repeated failed call; ineffective fallback
& Visible failure feedback is not followed by an appropriate response. \\
Evidence & Incomplete retrieval; source substitution; mismatched evidence
& Required evidence is retrievable but collected or associated incorrectly. \\
Synthesis & Deliverable omission; truncation; format or consistency failure
& Evidence is available, but the final response is incomplete or inconsistent. \\
\bottomrule
\end{tabularx}
\end{table}

A stored skill contains a trigger, procedure, stop condition, anti-patterns,
applicability metadata, and provenance. It may refer only to information visible
in the task and trajectory. The router requires the same failure type, prefers
an exact subtype match over a wildcard match, checks server compatibility, and
injects at most one skill. If no active skill applies, a new candidate is
distilled from the failed rollout and visible verifier evidence. It becomes
reusable only after a fresh source-task recovery passes the original verifier.

Figures~\ref{fig:ranked-segment-skill} and~\ref{fig:transient-read-skill} show
two representative recovery skills. The cards omit source-task values while
preserving the reusable trigger, procedure, stop condition, anti-patterns, and
scope.

\newtcolorbox{skillcard}[1]{enhanced,nobeforeafter,
  colback=white,colframe=promptblue!80!black,
  colbacktitle=promptblue!88!black,coltitle=white,
  fonttitle=\bfseries\small,fontupper=\footnotesize,
  title={#1},boxrule=0.6pt,arc=1.5pt,
  left=8pt,right=8pt,top=6pt,bottom=6pt,
  before upper={\setlength{\parindent}{0pt}\setlength{\parskip}{3pt}}}
\newcommand{\skillmeta}[1]{\colorbox{blue!7}{%
  \parbox{\dimexpr\linewidth-2\fboxsep\relax}{\footnotesize #1}}}
\newcommand{\skillstep}[2]{\par\noindent\hangindent=1.55em\hangafter=1%
  \textbf{#1.}\enspace #2}

\begin{figure}[H]
\centering
\begin{skillcard}{Synthesis skill: reconcile ranked top-$k$ and remaining segments}
\textbf{Trigger.} Use when grouped counts must be divided into a leading
top-$k$ segment and a remaining segment, with totals, percentages, ratios, or
averages reconciled in the final answer.

\textbf{Procedure.}
\skillstep{1}{Group by the target field, sort the complete ranked list, and
inspect all groups before selecting a split point.}
\skillstep{2}{If $k$ is not fixed by the request, infer a coherent leading
segment from the ranking and later references to the remaining groups; do not
default to top-3.}
\skillstep{3}{Compute the leading and remaining totals from the grouped results
returned for the current task.}
\skillstep{4}{Verify that the two segment totals reconcile with an independently
observed collection total.}
\skillstep{5}{Derive every requested comparison, including shares, ratios,
segment averages, and the range of the remaining groups.}
\skillstep{6}{Before answering, check that the selected collection, field
validation, ranked groups, segment labels, totals, comparisons, and conclusion
are all present.}
\skillstep{7}{If an earlier calculation used a different split, correct it
explicitly rather than leaving contradictory segment definitions.}

\textbf{Stop condition.} Stop when the selected segments are correct, their
totals match the independent count, and every requested comparison is stated.

\textbf{Avoid.} Defaulting to top-3; listing group counts without segment
totals; mixing different values of $k$; or drawing a conclusion before the
numeric reconciliation.

\skillmeta{\textbf{Scope:} Teacher-only grouped-data synthesis repair
\hfill \textbf{Status:} Active; source recovery passed}
\end{skillcard}
\caption{\textbf{Ranked-segment reconciliation skill.} The procedure partitions
ranked groups, reconciles their totals, and checks every requested comparison
without retaining source-task values.}
\label{fig:ranked-segment-skill}
\end{figure}

\begin{figure}[H]
\centering
\begin{skillcard}{Evidence skill: continue after a transient read rejection}
\textbf{Trigger.} Use when a read-only Notion request is rejected or temporarily
unavailable while the task still requires evidence from accessible pages.

\textbf{Procedure.}
\skillstep{1}{Retry the same read once when the error explicitly indicates a
transient or retryable failure.}
\skillstep{2}{After a database query succeeds, enumerate the visible candidate
page identifiers and titles before selecting pages to inspect.}
\skillstep{3}{For each selected page, request its child blocks and record the
returned results, pagination flag, and cursor.}
\skillstep{4}{If a child-block request is rejected, retry it once when the
failure is retryable; otherwise, do not claim that the page was inspected.}
\skillstep{5}{Continue with another visible candidate page until the required
successful inspections are complete or no candidate remains.}
\skillstep{6}{In the final answer, distinguish successful observations from
rejected attempts; only successful calls support claims about page content.}
\skillstep{7}{Ground database identity, page identity, and block structure in
their corresponding visible tool responses.}

\textbf{Stop condition.} Stop when all required pages have successful child-block
responses, or when every visible candidate has been attempted and the answer
explicitly limits its claims to successfully inspected pages.

\textbf{Avoid.} Treating the first rejection as permanent; claiming an empty
result after a rejected call; stopping after one page when several are required;
inventing page names or outcomes; or retrying a non-retryable failure indefinitely.

\skillmeta{\textbf{Scope:} Teacher-only Notion evidence collection
\hfill \textbf{Status:} Active; source recovery passed}
\end{skillcard}
\caption{\textbf{Complete transient-read recovery skill.} This active skill
continues evidence collection after a temporary tool rejection while preventing
failed calls from being presented as successful observations.}
\label{fig:transient-read-skill}
\end{figure}

\subsection{Efficiency selection and SFT export}
\label{app:trajectory_efficiency}
\label{app:trajectory_cleaning}

The tool-path and assistant-output audits produce the quantities used by AWR in
Equation~\ref{eq:avoidable-work-ratio}. Tool-error and AWR thresholds are set
per task rather than globally. Candidates that do not meet their task-specific
thresholds are rerun on the unchanged task in a fresh session, with the audit
findings supplied as teacher-only refinement guidance. If no additional
candidate meets both thresholds, the implementation selects among correct
candidates by the ordered key
$(e(\tau),\operatorname{AWR}(\tau),k)$, where $e(\tau)$ is the number of real
tool errors and $k$ is the generation index. The original correct trajectory is
included in this candidate set.

The SFT exporter retains the public request, model-visible reasoning, tool calls,
tool observations, and final answer. Routed skills, failure diagnoses, verifier
feedback, efficiency judgments, and other teacher-only metadata are removed.
The exporter also checks that the call/result sequence is complete and that
every invoked tool belongs to the task's recorded interface.

\section{Complete EvoMCP Procedure}
\label{app:algorithm}

Algorithm~\ref{alg:evomcp-round} gives the round-level procedure that connects
the two evolution stages described in Section~\ref{sec:method}.

\begin{algorithm}[H]
\caption{One round of \method}
\label{alg:evomcp-round}
\begin{algorithmic}[1]
\Require Student $\pi_t$; verified pool $\D_t$; skill pool $\mathcal{K}_t$; budget $B_t$
\Ensure Student $\pi_{t+1}$; verified pool $\D_{t+1}$; skill pool $\mathcal{K}_{t+1}$
\State Evaluate $\pi_t$ and summarize feedback signals $\mathcal{S}_t$
\State Initialize $\D_t^{\mathrm{train}}\gets\varnothing$ and $\mathcal{K}_{t+1}\gets\mathcal{K}_t$
\While{budget $B_t$ remains}
    \State Select a learning target and compatible verified seeds from $\D_t$
    \State Evolve the task instruction, environment, or both to obtain $x'=(q',E',V')$
    \If{$x'$ fails consistency, materialization, or verifier checks}
        \State reject $x'$ and continue
    \EndIf
    \State Execute a fresh teacher rollout $\tau_0$ on $x'$
    \If{$V'(\tau_0)=0$ and the failure is attributable to the teacher}
        \If{a compatible active recovery skill exists}
            \State route the existing skill
        \Else
            \State distill a source-grounded recovery skill
        \EndIf
        \State retry in a fresh session on the unchanged task
    \EndIf
    \If{no teacher trajectory passes $V'$}
        \State reject the candidate and continue
    \EndIf
    \State Assess correct candidates for tool errors and avoidable work
    \If{a candidate misses its task-specific thresholds and budget remains}
        \State rerun on the unchanged task using the audit findings as guidance
        \State add only verifier-passing refinements to the candidate set
    \EndIf
    \State Select $\tau^\star$ and remove teacher-only guidance and annotations
    \State Add $(x',\tau^\star)$ to $\D_t^{\mathrm{train}}$
    \State Update $\mathcal{K}_{t+1}$ with reusable source-validated skills
\EndWhile
\State $\pi_{t+1}\gets\operatorname{SFT}(\pi_t,\D_t^{\mathrm{train}})$
\State $\D_{t+1}\gets\D_t\cup\D_t^{\mathrm{train}}$
\end{algorithmic}
\end{algorithm}

\section{Pipeline Prompt Templates}
\label{app:prompts}

The following cards show instructions for curriculum planning, task construction,
task realization, failure attribution, recovery-skill distillation, and efficiency
assessment. The planner and builder cards quote excerpts of the code templates;
their task-specific inputs are supplied at runtime. Efficiency assessment uses two calls
because tool-path errors and avoidable assistant output have different inputs
and output schemas. Task-specific runtime records are summarized after each
static instruction.

\newcommand{\promptpayload}[1]{\textbf{Runtime input:} \emph{#1}.}

\subsection{Curriculum planning and task construction}
\label{app:prompt_task_construction}

\begin{figure}[H]
\centering
\begin{promptcard}{Prompt 1: Curriculum planner}
\textbf{SYSTEM (excerpts):}

Allocate a reusable curriculum from the Student diagnosis and COMPLETE seed
roster. Use \texttt{curriculum\_allocation} once. Return only
\texttt{purpose\_weights} and compact lanes; do not author task goals, fields,
values, verifier logic, solving steps or implementation plans.

For each seed group choose \texttt{runtime\_seed\_options},
\texttt{mutation\_axes} and \texttt{change\_strength}. The axes are control
coordinates, NOT a task/goal/family whitelist. The local Builder receives the
full original goal and implementation materials and chooses the concrete
source-grounded business change.

\textbf{USER:}

\promptpayload{Student capability diagnosis, complete seed roster, available
MCP catalogue, curriculum policy and delivery ledger}
\end{promptcard}
\caption{Excerpts from the curriculum-planner system prompt. The output assigns
learning directions and seeds to construction work orders.}
\label{fig:curriculum-planner-prompt}
\end{figure}

\begin{figure}[H]
\centering
\begin{promptcard}{Prompt 2: Task builder}
\textbf{SYSTEM (excerpts):}

Realize ONE curriculum work order as a source-grounded task specification.
The Controller selected runtime seeds, one mutation axis and a change strength;
it did NOT author a concrete goal. You own the local business change after
reading the supplied original goal and \texttt{implementation\_surface}.
Seeds are initial data/environment plus an example goal, not a mandatory final goal.

Public tasks state the business outcome/rules, never tool sequences, Student
weaknesses, grader internals or private answers. Private \texttt{semantic\_delta}
explains what actually changes.

Use \texttt{task\_specification} with \texttt{action=build} and
\texttt{task=\{operation,parameters\}}. The framework computes references,
writes files, packages the unchanged seed environment, and executes the same
verifier entry used by the harness.

\textbf{USER:}

\promptpayload{Work order and learning target, full seed request, indexed source
materials, operation APIs, sibling goals, and available MCP tools}
\end{promptcard}
\caption{Excerpts from the builder system prompt for requirement changes on
selected seed environments. Recorded work orders and outputs appear in the cases
in Section~\ref{app:task_evolution_cases}.}
\label{fig:task-builder-prompt}
\end{figure}

\subsection{Task realization}
\label{app:prompt_task_realization}

\begin{figure}[H]
\centering
\begin{promptcard}{Prompt 3: Task realization}
\textbf{SYSTEM:}

You realize a typed tool-use task as one natural user request. Rewrite the whole
request; never append or concatenate parent prompts. Preserve every target goal,
dependency, constraint, and required verbatim clause. Do not mention internal
graphs, verifiers, operators, benchmarks, or evolution terminology.

Return JSON only with \texttt{prompt}, \texttt{covered\_node\_ids}, and
\texttt{intent\_summary}.

\textbf{USER:}

\promptpayload{Typed task program, parent prompts, surface contract, and target
signal}
\end{promptcard}
\caption{Static instruction used to realize a structured task as a user request.}
\label{fig:task-realization-prompt}
\end{figure}

\subsection{Failure attribution}
\label{app:prompt_failure_attribution}

\begin{figure}[H]
\centering
\begin{promptcard}{Prompt 4: Failure attribution}
\textbf{SYSTEM:}

You are a control-plane component for an executable MCP self-evolution system.
Return one JSON object only. Do not invent executed evidence. Separate model
behavior from environment, harness, and verifier or task-contract failures.

\textbf{USER:}

Diagnose only the supplied failed teacher rollouts. Return
\texttt{primary\_attribution}, \texttt{failure\_atoms},
\texttt{capability\_contract}, \texttt{confidence}, \texttt{evidence},
\texttt{recommended\_route}, \texttt{affected\_attempt\_ids},
\texttt{failure\_type}, and \texttt{failure\_subtype}. Attribute model behavior
only when it is supported by the teacher trace and unchanged verifier.

\promptpayload{Public task, failed trace, verifier outcome, and visible
execution evidence}
\end{promptcard}
\caption{Static instruction used for teacher-failure attribution.}
\label{fig:failure-attribution-prompt}
\end{figure}

\subsection{Recovery-skill distillation}
\label{app:prompt_skill_distillation}

\begin{figure}[H]
\centering
\begin{promptcard}{Prompt 5: Recovery-skill distillation}
\textbf{SYSTEM:}

You are a control-plane component for an executable MCP self-evolution system.
Return one JSON object only and do not invent executed evidence.

\textbf{USER:}

Compare the failed teacher trajectory, the agent-visible request, tool responses,
final artifacts, and the failed verifier contract. Decide whether the failure is
\texttt{recoverable} or a \texttt{task\_contract\_mismatch}. If recoverable,
return one temporary repair skill with \texttt{skill\_id}, \texttt{trigger},
\texttt{procedure}, \texttt{stop\_condition}, \texttt{anti\_patterns}, and
\texttt{scope}. The procedure may improve planning, tool use, checking, or
recovery, but must not change the task, environment, verifier, expected answer,
or available tools. Ground every step in supplied evidence. Do not encode
verifier-only criteria, hidden reference answers, or information unavailable to
the teacher during execution. This is teacher-only generation context.

\promptpayload{Failed rollout, diagnosis, visible evidence, and compatible
skills}
\end{promptcard}
\caption{Static instruction used to distill a source-grounded recovery skill.}
\label{fig:skill-distillation-prompt}
\end{figure}

\subsection{Trajectory-efficiency assessment}
\label{app:prompt_efficiency_assessment}

\begin{figure}[H]
\centering
\begin{promptcard}{Prompt 6(a): Tool-path efficiency audit}
\textbf{SYSTEM:}

You are a strict trajectory-efficiency auditor. A redundant successful call is
wholly avoidable: it repeats already-known evidence or never contributes to a
required inference. Do not mark a call merely because another strategy was
possible. A real tool error means that the call failed, timed out, was rejected,
returned an explicit error, or used invalid arguments. Use only listed call IDs.

Return JSON with \texttt{redundant\_successful\_call\_ids},
\texttt{additional\_real\_error\_call\_ids}, and \texttt{notes}.

\textbf{USER:}

\promptpayload{Public task and ordered tool-call summaries}
\end{promptcard}
\caption{Static instruction used for tool-path efficiency assessment.}
\label{fig:tool-efficiency-prompt}
\end{figure}

\begin{figure}[H]
\centering
\begin{promptcard}{Prompt 6(b): Assistant-output efficiency audit}
\textbf{SYSTEM:}

For every assistant-output segment, estimate the fraction that is avoidable
without reducing correctness, evidence traceability, or task compliance.
Necessary planning, evidence interpretation, calculations, and final synthesis
are not waste. Repetition, abandoned narration, needless restatement, and
unrelated text are waste. Choose one value from
$\{0,0.05,0.10,0.20,0.30,0.40,0.50,0.75,1.0\}$ and return each supplied
segment identifier exactly once.

\textbf{USER:}

\promptpayload{Public task, tool-path summary, and ordered assistant segments}
\end{promptcard}
\caption{Static instruction used for output-efficiency assessment.}
\label{fig:output-efficiency-prompt}
\end{figure}

\section{Qualitative Cases}
\label{app:cases}

The following figures show recorded requirement changes, environment variants,
and a requirement change on an evolved environment. The task cases connect the
changed request or state to the corresponding verification. The trajectory figure
abridges a recorded recovery while preserving its tool calls, intermediate
values, and verifier-grounded failure atoms.

\newtcolorbox{evolutioncase}[2][]{enhanced,nobeforeafter,
  colback=blue!2,colframe=promptblue,colbacktitle=promptblue,
  coltitle=white,fonttitle=\bfseries\small,fontupper=\footnotesize,
  title={#2},boxrule=0.6pt,arc=1.5pt,valign=top,
  left=6pt,right=6pt,top=6pt,bottom=6pt,
  before upper={\raggedright\setlength{\parindent}{0pt}\setlength{\parskip}{4pt}},#1}
\newcommand{\caseblue}[1]{\colorbox{blue!12}{\parbox{\dimexpr\linewidth-2\fboxsep\relax}{#1}}}
\newcommand{\casered}[1]{\colorbox{red!12}{\parbox{\dimexpr\linewidth-2\fboxsep\relax}{#1}}}
\newcommand{\casegray}[1]{\colorbox{black!5}{\parbox{\dimexpr\linewidth-2\fboxsep\relax}{#1}}}
\newcommand{\casemetric}[1]{\colorbox{black!5}{\parbox{\dimexpr\linewidth-2\fboxsep\relax}{\centering\scriptsize #1}}}

\subsection{Task-evolution cases}
\label{app:task_evolution_cases}

\begin{figure}[H]
\centering
\textbf{Case A: replacing a single-run goal with checkpoint-wise reconciliation}\par\smallskip
\casegray{\footnotesize\raggedright\textbf{Feedback and planner assignment.} The capability report
identifies incomplete collection of experiment and research objects and incomplete
structured deliverables. The recorded assignment selects the W\&B seed with
\texttt{purpose=explore}, \texttt{mutation\_axis=goal\_replacement}, and
\texttt{change\_strength=structural}.}
\smallskip
\begin{minipage}[t]{0.475\linewidth}
\begin{evolutioncase}[height=61mm]{Before: seed requirement}
\textbf{Source.} A W\&B project with five completed runs, each containing
500 logged steps.

\caseblue{\textbf{Goal.} Select the run with the smallest final response length.
Record that run's entropy loss, clip ratio, and mean response length at
checkpoint zero and subsequent 100-step checkpoints.}

\textbf{Output.} \texttt{shortest\_length\_experiment.csv}.

\textbf{Selection.} One final winner supplies the metrics across checkpoints.
\end{evolutioncase}
\end{minipage}
\hfill
\begin{minipage}[t]{0.475\linewidth}
\begin{evolutioncase}[height=61mm]{After: evolved requirement}
\textbf{Source.} The same project and run histories.

\caseblue{\textbf{Goal.} At each checkpoint, compare all five runs and select
the one with the smallest mean response length. Record its metrics and whether
it is also the final-step winner. Include steps 0, 100, 200, 300, 400, and 499.}

\textbf{Output.} \texttt{checkpoint\_leader\_\allowbreak{}reconciliation.csv}.

\textbf{Selection.} The leading run is determined separately at every checkpoint.
\end{evolutioncase}
\end{minipage}
\smallskip
\casegray{\footnotesize\raggedright\textbf{Builder and validation.} The builder patches the seed
evaluator for the new table and selection rule. It accepts two positive cases and
rejects seven negative cases, including the old single-winner table and an omitted
final checkpoint. A teacher rollout then passes the native verifier.}
\caption{\textbf{Requirement evolution.} The planner supplies the learning
direction; the builder chooses a concrete change supported by the seed data.
The recorded assignment is exploratory. Both the output requirement and its
verifier change, while the source environment is reused.}
\label{fig:task-evolution-case}
\end{figure}

\begin{figure}[H]
\centering
\textbf{Case B: changing memory facts and the live form layout}\par\smallskip
\casegray{\footnotesize\raggedright\textbf{Shared request.} Complete the welcome-party form for two
people using their stored memory. Both attend the full day; unstated information
defaults to negative. If a person has multiple responses, the latest one counts.}
\smallskip
\begin{minipage}[t]{0.475\linewidth}
\begin{evolutioncase}[height=57mm]{Environment instance: s18}
\textbf{Memory facts.}

\caseblue{Elena Brooks S0018: No Seafood.\\
Talia Nguyen S0018: None.}

\textbf{Form state.} Eleven questions in a form created for this instance.
\texttt{Name} is question 3; \texttt{Dietary Restrictions} is question 4.

\textbf{Dietary option order.} None; No Seafood; No Spicy Food; Kosher; Vegan.

\textbf{Expected responses.} Bound to the s18 people, facts, and form.
\end{evolutioncase}
\end{minipage}
\hfill
\begin{minipage}[t]{0.475\linewidth}
\begin{evolutioncase}[height=57mm]{Environment instance: s21}
\textbf{Memory facts.}

\caseblue{Noah Patel S0021: Vegan.\\
Priya Martin S0021: No Seafood.}

\textbf{Form state.} Eleven questions in a separately initialized form.
\texttt{Name} is question 7; \texttt{Dietary Restrictions} is question 11.

\textbf{Dietary option order.} No Spicy Food; None; No Seafood; Vegan; Kosher.

\textbf{Expected responses.} Bound to the s21 people, facts, and form.
\end{evolutioncase}
\end{minipage}
\smallskip
\casegray{\footnotesize\raggedright\textbf{Validation.} The request and MCP tool interfaces are the same.
The evaluator retains exact scalar matching, checkbox-set matching, and the
latest-response rule. Local checks reject wrong choices, incomplete attendance,
and stale responses. Teacher rollouts pass the native verifier in both instances.}
\caption{\textbf{Environment variation.} Recorded instances from the paired
environment evaluation change observable facts and layout while retaining the
capability goal. Expected response values and the target form are rebound to each
instance. The held-out s21 instance and its teacher trajectory are excluded from
SFT.}
\label{fig:environment-evolution-case}
\end{figure}

\begin{figure}[H]
\centering
\textbf{Case C: requirement evolution on an evolved Canvas environment}\par\smallskip
\casegray{\footnotesize\raggedright\textbf{Environment preparation.} The seed varies course names,
teacher assignments, meeting times, schedule order, and syllabus-file mappings.
Its emailed schedule contains 27 assigned courses taught by 11 instructors.
The goal change below operates on this initialized environment.}
\smallskip
\casegray{\footnotesize\raggedright\textbf{Feedback and planner assignment.} The capability report
identifies incomplete Canvas side effects and incomplete entity coverage.
The planner selects \texttt{purpose=remediate},
\texttt{mutation\_axis=grouping}, and \texttt{change\_strength=structural}.}
\smallskip
\begin{minipage}[t]{0.475\linewidth}
\begin{evolutioncase}[height=65mm]{Before: per-course announcements}
\textbf{Example instructor.} Rebecca Richardson teaches:

\casegray{Modern Poetry --- Cedar Term 68;\\
American Literature --- Alder Term 68;\\
Developmental Psychology --- Harbor Term 68.}

\caseblue{\textbf{Requirement.} Publish an announcement in each assigned
course stating its instructor and class time.}

\textbf{Other obligations.} Create each course, enroll its instructor, upload
its syllabus, and publish the course.
\end{evolutioncase}
\end{minipage}
\hfill
\begin{minipage}[t]{0.475\linewidth}
\begin{evolutioncase}[height=65mm]{After: announcements grouped by instructor}
\textbf{Same schedule and course resources.}

\caseblue{\textbf{Requirement.} For each instructor, publish the consolidated
notice in the first course in schedule order. Name that instructor and every
course in the group using the full scheduled titles. Other courses in the group
must not carry a notice naming that instructor.}

\textbf{Example destination.} Modern Poetry --- Cedar Term 68 carries the notice
for Rebecca Richardson's three courses.

\textbf{Preserved.} Course creation, teacher enrollment, syllabus upload, and
course publication.
\end{evolutioncase}
\end{minipage}
\smallskip
\casegray{\footnotesize\raggedright\textbf{Builder and validation.} The builder groups by instructor,
chooses the first course, and updates the announcement checks. It accepts one
positive case and rejects six negative cases, covering missing group members, a notice in the
wrong course, and missing retained side effects. The teacher passes the native
verifier.}
\caption{\textbf{Combining environment and requirement changes.} Environment
variation first supplies the executable seed; feedback-guided requirement
evolution then changes announcement routing on that seed. The final task uses
11 instructor-group destinations while preserving the remaining requirements for
all 27 courses.}
\label{fig:combined-task-evolution-case}
\end{figure}

\subsection{Trajectory-evolution case}
\label{app:trajectory_evolution_case}

\begin{figure}[H]
\centering
{\small\bfseries Same task; skill-guided recovery proceeds from left to right.}\par\smallskip
\begin{minipage}[t]{0.475\linewidth}
\begin{evolutioncase}[height=84mm]{Before: failed trajectory \hfill \textcolor{red!25!white}{FAIL}}
\casegray{\textbf{1. Locate data}\quad
\texttt{list-collections} $\rightarrow$ six schema inspections}

\casered{\textbf{2. Aggregate}\quad
\texttt{group(Game Genre, count)}\\
Counts were collected without the required sales evidence.}

\casered{\textbf{3. Select samples}\quad
\texttt{find(..., limit=1)}\\
The returned documents did not match the required samples.}

\casered{\textbf{4. Reconcile groups}\quad
Top 1: \textbf{423}\quad vs.\quad remainder: \textbf{2,371}\\
The required top-four--tail comparison was not performed.}

\casered{\textbf{5. Final evidence}\quad
Average Daily Sales and the tail's sales contribution were absent.}

\casemetric{\textbf{10} turns \quad $\vert$ \quad \textbf{31} tool calls \quad $\vert$ \quad
claim coverage: \textbf{0.50}}
\end{evolutioncase}
\end{minipage}
\hfill
\begin{minipage}[t]{0.475\linewidth}
\begin{evolutioncase}[height=84mm]{After: recovered trajectory \hfill \textcolor{green!55!black}{PASS}}
\casegray{\textbf{1. Locate data}\quad
\texttt{list-collections} $\rightarrow$ four focused schema inspections}

\caseblue{\textbf{2. Aggregate}\quad
\texttt{group(Game Genre, count, sales)}\\
Counts, average sales, and total sales were collected together.}

\caseblue{\textbf{3. Select samples}\quad
\texttt{find(..., sort=\_id)} $+$ count checks\\
The deterministic samples agreed with the grouped counts.}

\caseblue{\textbf{4. Reconcile groups}\quad
Top 4: \textbf{1,666}\quad vs.\quad tail: \textbf{1,128}\\
The two groups account for all \textbf{2,794} documents.}

\caseblue{\textbf{5. Final evidence}\quad
Tail sales: \textbf{5,792.16 / 13,663.49} (\textbf{42.4\%}).}

\casemetric{\textbf{7} turns \quad $\vert$ \quad \textbf{20} tool calls \quad $\vert$ \quad
claim coverage: \textbf{1.00}}
\end{evolutioncase}
\end{minipage}
\caption{\textbf{Trajectory-evolution case.} The two columns show an abridged
trace for the same task. Red rows identify verifier-grounded failures; blue rows
show the corresponding skill-guided corrections. The tool calls, intermediate
values, and footer statistics come from the complete recorded executions. Claim
coverage is the score reported by the task verifier.}
\label{fig:trajectory-evolution-case}
\end{figure}

\clearpage
\begingroup
\section{Additional Experimental Analyses}

\subsection{Benchmark feedback and evaluation}
\label{app:benchmark_feedback}

The task sets from Toolathlon, MCPMark, and MCP-Atlas supply student rollout feedback and are also used to score the checkpoints in Tables~\ref{tab:model-comparison} and~\ref{tab:training-comparison}. After each student update, rollouts and task outcomes on these benchmark tasks inform the learning targets and seed-task selection for the next evolution round. We evaluate successive checkpoints on the same task sets. The reported scores therefore measure improvement under benchmark-guided feedback across rounds.

\subsection{Student progress on the feedback pool}
\label{app:student_progress}

We align the existing MCP-Atlas diagnostic records by task ID. The three checkpoints share 310 tasks with identical queries and reference claims. Their Pass@1 scores are shown in Table~\ref{tab:diagnostic-rounds}. The initial checkpoint is the student before the reported evolution rounds, rather than the seed-only SFT baseline.

\begin{table}[H]
\caption{\textbf{Student performance on the common diagnostic tasks.}}
\label{tab:diagnostic-rounds}
\centering
\begin{tabularx}{0.84\linewidth}{@{}Yrrr@{}}
\toprule
\textbf{Checkpoint} & \textbf{Tasks} & \textbf{Passed} & \textbf{Pass@1 (\%)} \\
\midrule
Initial & 310 & 196 & 63.23 \\
\method{} R1 & 310 & 221 & 71.29 \\
\method{} R2 & 310 & 228 & 73.55 \\
\bottomrule
\end{tabularx}
\end{table}

Eight tasks lose Lara translation tools after the initial evaluation. For the paired analysis, we use the remaining 302 tasks, whose queries, reference claims, and enabled tool-name lists are unchanged. Table~\ref{tab:diagnostic-transitions} groups tasks by the initial outcome, so that improvement on failures and retention of prior successes are both visible.

\begin{table}[H]
\caption{\textbf{Resolving initial failures and retaining initial successes.} Percentages use the number of tasks in each initial group.}
\label{tab:diagnostic-transitions}
\centering
\small
\begin{tabularx}{\linewidth}{@{}Yrrr@{}}
\toprule
\textbf{Initial outcome} & \textbf{Tasks} & \textbf{R1 passed} & \textbf{R2 passed} \\
\midrule
Failed & 113 & 60 (53.10\%) & 65 (57.52\%) \\
Passed & 189 & 154 (81.48\%) & 155 (82.01\%) \\
\midrule
All & 302 & 214 (70.86\%) & 220 (72.85\%) \\
\bottomrule
\end{tabularx}
\end{table}

Relative to the initial student's 189/302 passing tasks, R2 gains 65 passes and loses 34, for a net increase of 31 tasks or 10.26 percentage points. These diagnostic outcomes describe changes in observed failures. The feedback ablation in Table~\ref{tab:training-comparison} separately tests the effect of feedback-guided curriculum planning.

\paragraph{Variation across reference workflows.}
We also group the same 302 tasks using fixed reference trajectories. Calculation-related workflows contain a calculator, code executor, or MongoDB aggregation call. Service counts describe the services used in the reference execution. Table~\ref{tab:diagnostic-workflows} shows that changes differ across groups. Calculation-related performance peaks at R1, while tasks whose references use at least three services continue to improve at R2.

\begin{table}[H]
\caption{\textbf{Diagnostic Pass@1 by reference workflow.}}
\label{tab:diagnostic-workflows}
\centering
\small
\begin{tabularx}{\linewidth}{@{}Yrrrr@{}}
\toprule
\textbf{Reference workflow} & \textbf{N} & \textbf{Initial} & \textbf{R1} & \textbf{R2} \\
\midrule
Calculation-related & 99 & 56.57 & 73.74 & 70.71 \\
Other workflows & 203 & 65.52 & 69.46 & 73.89 \\
\midrule
One service & 5 & 100.00 & 100.00 & 80.00 \\
Two services & 209 & 64.11 & 75.12 & 76.08 \\
At least three services & 88 & 56.82 & 59.09 & 64.77 \\
\bottomrule
\end{tabularx}
\end{table}

\clearpage
\subsection{A behavior-level correction}
\label{app:student_case}

One diagnostic task requests apartments built after 2010, subject to additional housing constraints, and asks for the cheapest apartment and the mean rent. Table~\ref{tab:student-boundary-case} traces a concrete correction in the student's tool arguments and answer.

\begin{table}[H]
\caption{\textbf{Correcting a strict filter boundary.} All entries come from recorded tool calls and answers for the same task.}
\label{tab:student-boundary-case}
\centering
\small
\begin{tabularx}{\linewidth}{@{}Yccc@{}}
\toprule
\textbf{Observed behavior} & \textbf{Initial} & \textbf{R1} & \textbf{R2} \\
\midrule
YearBuilt predicate & $\geq 2010$ & $>2010$ & $>2010$ \\
Selected apartments & 12 & 11 & 11 \\
Reported mean rent & 3,293.583 & 3,387.182 & 3,387.182 \\
Recorded claim coverage & 0.167 & 1.000 & 1.000 \\
\bottomrule
\end{tabularx}
\end{table}

The initial student includes apartments built in 2010. Both evolved checkpoints apply the strict boundary and compute the mean over the resulting 11 apartments. This case identifies a specific learning target: translating a verbal boundary into the correct tool predicate and carrying the selected set consistently into aggregation. R2 still reports an incorrect auxiliary sum in its final answer, despite the correct mean and a passing judge verdict.

\subsection{Task-evolution measurements}
\label{app:task_metrics}

The following protocol specifies the paired comparison in Table~\ref{tab:mechanisms}(a).

\paragraph{Sample and round.}
The sample consists of 2,400 single-parent seed--evolved task pairs from task acquisition before the first student update, denoted R1. Eligible evolved tasks have complete lineage and pass task-validity checks. Pairs are sampled uniformly without replacement using random seed 42 before filtering on teacher success. Each evolved-task ID appears once, although a seed may support different targets or descendants. Multi-seed constructions are outside this paired comparison. The target is the planner's recorded pre-generation target, shared by both members and kept unchanged during scoring. The pair record retains task versions, the target, and its retain, remediate, or explore purpose. Each pair has equal weight.

\paragraph{Target realization.}
A task realizes its target when its requirements necessarily exercise the selected behavior, its observable environment provides the information or resources needed for that behavior, and its verifier checks the corresponding requirement. The judgment must identify evidence for all three conditions in the task instruction, environment resources, and verifier code or tests. A target label, a changed instruction, or a successful rollout alone is insufficient. The seed and evolved task receive separate judgments against the same target. Both can be positive when the seed already exercises that behavior.

\paragraph{Behavior change.}
An evolved task is positive when the comparison identifies a changed necessary action, dependency, aggregation rule, or decision conditioned on observable state. For an environment transformation, the evidence must explain how changed facts alter the required action, lookup, or tool arguments. Wording changes and incidental differences between sampled trajectories do not qualify. The comparison records the required behavior before and after evolution and cites the task evidence establishing the difference. Behavior change is reported only for evolved tasks, with seeds serving as the reference.

\paragraph{Judge and consistency check.}
GLM-5.2 is the fixed audit judge, with temperature 0 and an 8,192-token output limit. Its input contains the unchanged learning target and both task packages, including instructions, relevant environment resources, verifier checks, and verifier-test evidence. Generator rationales, student and teacher outcomes, and round or source labels are withheld. Two separate calls present the packages as A and B in opposite orders. Each call returns target-realization labels for A and B, a behavior-change label, and source locations supporting every label. Labels are positive, negative, or insufficient evidence. A positive score requires both calls to agree after restoring task identities and to supply the evidence required above. All other cases score zero, with disagreements and insufficient-evidence cases retained in the audit records and denominator. These are agreement-confirmed model judgments, not human annotations.

\begin{quote}
\small
\textbf{Audit instruction.} Assess both task packages against the supplied learning target. For each package, identify the necessary task behavior, the accessible information or resources that support it, and the verifier check that tests it. Label target realization positive only when all three are supported. Then determine whether solving the two tasks requires a substantive difference in actions, dependencies, aggregation, or decisions based on observable state. Cite the before--after requirements and their source locations. Ignore wording-only changes, generation claims, and presumed model difficulty. Return a label and evidence for each judgment. Use insufficient evidence whenever a required link cannot be established.
\end{quote}

\paragraph{Student and teacher execution.}
The student is the initial Qwen3.6-27B checkpoint, R0, on both sides. The teacher is Qwen3.8-Max, using the same endpoint revision for all paired runs. Each actor receives one rollout per task from a fresh initial state with the standard task-execution prompt and declared tools. The paired protocol uses the Toolathlon single-turn harness with a 200-step agent limit, a configured 50-turn outer limit, temperature 0, maximum reasoning effort where supported, and a 262,144-token completion ceiling per model request. These limits match across the seed and evolved sides. Teacher rollouts receive no recovery skills or refinement guidance. Each task's own verifier determines success. A completed failing rollout, incorrect tool use, or exhaustion of the execution budget counts as failure. Infrastructure faults trigger a fresh-state rerun under unchanged settings and do not count as model failures. Execution rates are computed only after all 2,400 pairs have determinate outcomes. Existing seed demonstrations do not substitute for these matched executions.

\paragraph{Aggregation and interpretation.}
All percentages in Table~\ref{tab:mechanisms}(a) use the full set of 2,400 pairs. Target realization counts confirmed-positive tasks on each side, and behavior change counts confirmed-positive pairs. Student failure and teacher success count the respective verifier outcomes for R0 and the teacher. The difference column reports evolved-minus-seed differences in percentage points. The student failure rate has no universally preferred direction: retain targets seek to preserve an established behavior, while remediate and explore targets can expose unmet demands. Target realization can remain unchanged when evolution varies conditions for behavior already required by the seed.

\paragraph{Trajectory measurements.}
The first-round (R1) comparison in Table~\ref{tab:mechanisms}(b) uses 2,400 paired teacher demonstrations before and after refinement. Each pair shares the same task and verifier, with verifier-passing demonstrations on both sides. For each condition, we average tool-call errors, AWR, and generated output tokens across demonstrations. Errors and AWR follow the trajectory audit.

\paragraph{Round-level measurements.}
The task-validity, trajectory-acceptance, and training-pair-admission rates in Figure~\ref{fig:evolution-dynamics}(b) are computed from the actual evolution records in each round, using the observed task counts. Trajectory acceptance includes verifier-passing teacher demonstrations obtained through failure recovery. For Figure~\ref{fig:evolution-dynamics}(c), turns and tool calls are measured from teacher trajectories, with each measure normalized by its median at R0.

\paragraph{Matched recovery comparison.}
Figure~\ref{fig:evolution-dynamics}(d) compares unguided retries and skill-guided recovery on model-behavior failures eligible for retry in each round. The two conditions use the same eligible failed tasks and a matched teacher-rollout budget. The recovered-task share is the number of these failures that yield a verifier-passing demonstration within that budget, divided by the total number of tasks in the corresponding round.

\endgroup

\end{document}